\documentclass[11pt,a4paper]{article}

\usepackage[british]{babel}
\usepackage[T1]{fontenc}
\usepackage{csquotes}

\usepackage[a4paper,top=2cm,bottom=2cm,left=2.5cm,right=2.5cm,marginparwidth=1.75cm]{geometry}

\usepackage[
    style=numeric,
    backend=biber,
    natbib=true,
    sorting=nyt,
    maxnames=1,            
    minnames=1,            
    maxbibnames=10,         
    minbibnames=10,        
    doi=true,          
    url=true,         
    eprint=true     
]{biblatex}

\usepackage{amsmath}
\usepackage{amssymb}
\usepackage{amsfonts}
\usepackage{mathrsfs}

\usepackage{graphicx}
\usepackage{subcaption}
\usepackage{float}
\graphicspath{{figures/}}

\usepackage{array}
\usepackage{booktabs}
\usepackage{tabularray}
\usepackage{threeparttable}
\usepackage{diagbox}
\usepackage{varwidth}
\newcolumntype{M}{>{\begin{varwidth}{4cm}}l<{\end{varwidth}}}

\usepackage{caption}
\usepackage[colorlinks=true, allcolors=blue]{hyperref}

\usepackage[title]{appendix}

\usepackage{enumitem}
\usepackage{pdflscape}

\usepackage{setspace}
\usepackage{titlesec}
\titleformat{\section}
    {\normalfont\Large\bfseries}{\thesection.}{1em}{}

\usepackage{authblk}

\usepackage{chngcntr}

\let\oldFootnote\footnote
\newcommand\nextToken\relax
\renewcommand\footnote[1]{%
    \oldFootnote{#1}\futurelet\nextToken\isFootnote}
\newcommand\isFootnote{%
    \ifx\footnote\nextToken\textsuperscript{,}\fi}



\title{\textbf{Doing More with Less: A Survey on Routing Strategies for Resource Optimisation in Large Language Model-Based Systems}}

\author[1, 2, *]{Clovis Varangot-Reille}
\author[1]{Christophe Bouvard}
\author[2]{Antoine Gourru}
\author[1]{Mathieu Ciancone}
\author[3]{Marion Schaeffer}
\author[2]{François Jacquenet}
\affil[1]{\small Wikit, Lyon, France}
\affil[2]{\small Laboratoire Hubert Curien, UMR CNRS 5516, Saint-Etienne, France}
\affil[3]{\small INSA Rouen Normandie, Rouen, France}
\affil[*]{\small \textit{Corresponding author: Clovis Varangot-Reille,}  {\fontfamily{pcr}\selectfont\texttt{clovis.varangot\{wikit.ai\}}}}

\begin{document}
\definecolor{codegreen}{rgb}{0,0.6,0}
\definecolor{codegray}{rgb}{0.5,0.5,0.5}
\definecolor{codepurple}{rgb}{0.58,0,0.82}
\definecolor{backcolour}{rgb}{0.95,0.95,0.92}
\date{}
\maketitle

\begin{abstract}

Large Language Model (LLM)-based systems, i.e. interconnected elements that include an LLM as a central component, such as conversational agents, are usually designed with monolithic, static architectures that rely on a single, general-purpose LLM to handle all user queries. However, these systems may be inefficient as different queries may require different levels of reasoning, domain knowledge or pre-processing. While generalist LLMs (e.g. GPT-4o, Claude-Sonnet) perform well across a wide range of tasks, they may incur significant financial, energy and computational costs. These costs may be disproportionate for simpler queries, resulting in unnecessary resource utilisation. A routing mechanism can therefore be employed to route queries to more appropriate components, such as smaller or specialised models, thereby improving efficiency and optimising resource consumption. This survey aims to provide a comprehensive overview of routing strategies in LLM-based systems. Specifically, it reviews when, why, and how routing should be integrated into LLM pipelines to improve efficiency, scalability, and performance. We define the objectives to optimise, such as cost minimisation and performance maximisation, and discuss the timing of routing within the LLM workflow, whether it occurs before or after generation. We also detail the various implementation strategies, including similarity-based, supervised, reinforcement learning-based, and generative methods. Practical considerations such as industrial applications and current limitations are also examined, like standardising routing experiments, accounting for non-financial costs, and designing adaptive strategies.     Routing offers a practical way to improve the efficiency of LLM-based systems. By formalising routing as a performance–cost optimisation problem, this survey provides tools and directions to guide future research and development of adaptive low-cost LLM-based systems.

\end{abstract}



\section{Background}\label{background}
In Large Language Model (LLM)-based systems, a router is a component that routes an item (e.g. a user query) to the most suitable element in a pool of candidates to carry out a task, thereby optimising the consumption of resources. \\
This concept differs from the Mixture-of-Experts (MoE) framework \cite{jacobs1991}, where multiple "experts" learn to handle different parts of the input space. In traditional MoE architectures, an internal gating mechanism, generally a linear layer with a softmax activation, selects a subset of expert subnetworks to process each input \cite{fedus2022, Cai_2025}. In contrast, the router: (i) operates externally and independently of the LLM latent forward process; (ii) operates under budget constraints when accessing external candidates; and finally, (iii) routing candidates are not limited to neural networks, as in traditional MoE, but can also include workflows, data sources, or other system components. \\ 

In a more formal way, a router is defined as follows. Given a set of $n$ models, and more generally routing candidates, $\mathcal{M} = \{M_1,...,M_n\}$, for a given query $q$, the router function $\mathcal{R}$ aims to maximise the scoring function $s$ (e.g. \emph{accuracy}) while adhering to a budget constraint $B$: 
 \begin{equation}\label{eq1}
 \begin{aligned}
    \mathcal{R}_{\mathcal{M}}(q) =&\arg\max_{M \in \mathcal{M}} s(q,M)\\
   &\text{s.t. } C_{M}(q) \le B
 \end{aligned}
 \end{equation}

where $C_{M}$ is the cost (e.g., $\$/token$) to call the model $M$ for a query $q$ and $B$ is the user budget. The budget could be the amount of resources available.\\

This article focuses on one of the most common applications of LLM-based systems: \emph{conversational agents} \cite{dam2024}. Conversational agents respond to user queries by simulating human conversations \cite{caldarini2022, lin2023}. They process users queries and provide relevant answers \cite{caldarini2022}. The Retrieval-Augmented Generation (RAG) architecture improves the relevance of the responses by implementing an information retrieval step to the process \cite{Lewis2020rag, gao2023}, retrieving content from a knowledge base to include alongside the user query in the model prompt. Even compared to fine-tuning, this was shown to reduce hallucinations \cite{bouvard}.

A straightforward optimisation of conversational agent architectures is selecting the appropriate pre-trained LLM to answer a query. Models differ significantly in parameter count (e.g., \textit{Llama-3.2-3B} vs. \textit{Llama-3.1-405B}) and domain specialisation (e.g., \textit{mathstral-7B-v0.1}\footnote{\href{https://mistral.ai/fr/news/mathstral/}{MistralAI's Mathstral}} for mathematical tasks or \textit{Med-Palm}\footnote{\href{https://sites.research.google/med-palm/}{Google's Med-Palm}} for medical-related tasks). Nevertheless, most systems still rely on a single generalist LLM to handle all queries.
Routing each query to the most efficient LLM can improve cost-efficiency and response quality by leveraging domain-specific expertise. This prevents the unnecessary use of oversized models, models with insufficient reasoning ability, or models lacking the knowledge required for a particular task. \\

Through this work, we conduct a state-of-the-art of the routing frameworks implemented in LLM-based systems. Previous works have addressed routing within broader surveys on LLM inference optimization \cite{wang2024, chen2025harnessingmultiplelargelanguage, chen2025surveycollaborativemechanismslarge}, while \citet{behera2025efficientmultillminferencecharacterization} provided an overview of routing frameworks with limited strategy coverage. Our survey offers an in-depth analysis, defines the fundamental components of routing frameworks and introduces a taxonomy that distinguishes pre-generation routing (predicting candidate performance) from post-generation or cascade routing (evaluating a candidate based its output). We provide an extensive coverage of the routing literature, addressing three core questions:

\begin{itemize}
\item \textit{Q1: What should the routing optimise?}
\item \textit{Q2: When should routing take place?}
\item \textit{Q3: How routing should be implemented?}
\end{itemize}

We also examine the key limitations and challenges of LLM-based routing systems, such as poor generalisation to new  LLM candidates, a lack of standardisation of the routing evaluation and limited diversity in routing tasks. We discuss potential solutions to these issues. We exclude approaches that select the best answer from the answers generated by potential routing candidates \cite{guha2024, si2023} or ensemble methods that merge outputs from multiple models \cite{jiang2023, wang2024fusing, hu2024_der}. While these strategies can be effective, they prioritise performance over cost efficiency and thus fall outside our definition of routing. \\

This survey is structured into several sections that focus on the critical aspects of routing. We begin by describing the essential elements of routing. Next, we examine the pipeline stage at which routing is implemented in the literature. Finally, we detail and classify routing strategies according to the frameworks used. We discuss these strategies considering industrial practices and highlight the key challenges that the field must address.

\section{Q1: What should the routing optimise?}\label{which-section}

The primary goal of an efficient routing system is to minimise unnecessary resource consumption and maximise performance by using the most suitable model or system component for the task. In essence, efficient routing aims to optimise the trade-off between performance and cost.

\subsection{A performance metric to maximise}\label{performance-section}

A key objective of an efficient router function is to maximise a scoring function that evaluates a model’s ability to generate accurate answers, as described in Equation \ref{eq1}.

There are several possible scoring function to maximise. The evaluation process in a traditional supervised learning framework involves comparing the generated answers with the ground truth. In cases where data lack annotation, including a human evaluator in the evaluation loop enables to assess whether the response is factually correct, in the expected format, and consistent with the expected ground truth \cite{chang2024acm}. However, evaluating thousands of queries demands substantial time and intensive human effort, affecting scalability. In addition, subjective bias might appear while evaluating, such as lack of expertise or preferences. More straightforward strategies include exact matching, partial matching, ROUGE score \cite{lin-2004-rouge}, or even semantic similarity to the ground truth \cite{chang2024acm, zhang2024}. However, these metrics often fall short in evaluating factual accuracy.

To address this, prior research has explored automatic evaluation methods using LLMs. These involve embedding both the rating criteria and the generated response within the prompt \cite{chiang2023}. While promising, their alignment with human judgment remains uncertain, and outcomes are sensitive to instruction clarity \cite{wei2024}. Many frameworks, such as RAGAS \cite{es2023}, have been proposed.\\
Preference learning, based on the preferences of human reviewers, can be used to assess the quality of responses in addition to traditional methods \cite{jiang2024pref}. Data related to preferences are typically represented as \( A \prec_x B \), indicating that for a given query \( x \), the user prefers the generated answer \( A \) over the alternative \( B \) \cite{Fürnkranz2016}.

\subsection{A cost to minimise}\label{cost-section}
Most existing LLM-based systems depend on API calls to proprietary models, such as those provided for example by OpenAI, Google, Anthropic, Perplexity, xAI, etc. In this context, the primary cost a router seeks to minimise is the price per token. To accurately estimate the total cost of running a pipeline, one must account for all pre-trained models invoked during the process, including both LLMs and embedding models in RAG settings. Other production-related costs, such as latency or computational costs, can also be minimised \cite{irugalbandara2024}. Latency is the delay between the user request and the pipeline response. Effective routing could reduce the average latency by routing simple user queries to LLMs that require fewer computing resources.\\
We can also consider minimising the environmental cost of an LLM-based system. As LLMs increase in usage and scale, their power and computing requirements grow significantly \cite{Luccioni2024}, growing the ecological impacts associated with LLM-based systems. This impact is often measured considering the application's energy consumption (kWh) or global warming potential (kgCO2eq). Various tools\footnote{\href{https://ecologits.ai/latest/}{Ecologits}, \href{https://github.com/mlco2/codecarbon}{CodeCarbon}, \href{https://mlco2.github.io/impact/}{MLCO2}, \href{https://github.com/Boavizta/boaviztapi}{Boavizta}} have been proposed to estimate this environmental footprint.

\section{Q2: When should routing take place?}\label{when-section}
We identify two key stages in the pipeline where routing may take place: \emph{pre-generation routing}, presented in Figure \ref{fig:fig1}, and \emph{post-generation routing} (also known as \emph{cascade routing}), presented in Figure \ref{fig:fig2}. Pre-generation routing occurs before generating a response to the user query, while post-generation routing occurs after generating the response.

\subsection{Routing as a Pre-generation Step}\label{Routing as a Pre-generation Step}

\begin{figure*}[!htbp]
  \centering
  \makebox[\textwidth][c]{%
    \includegraphics[width=1.2\textwidth]{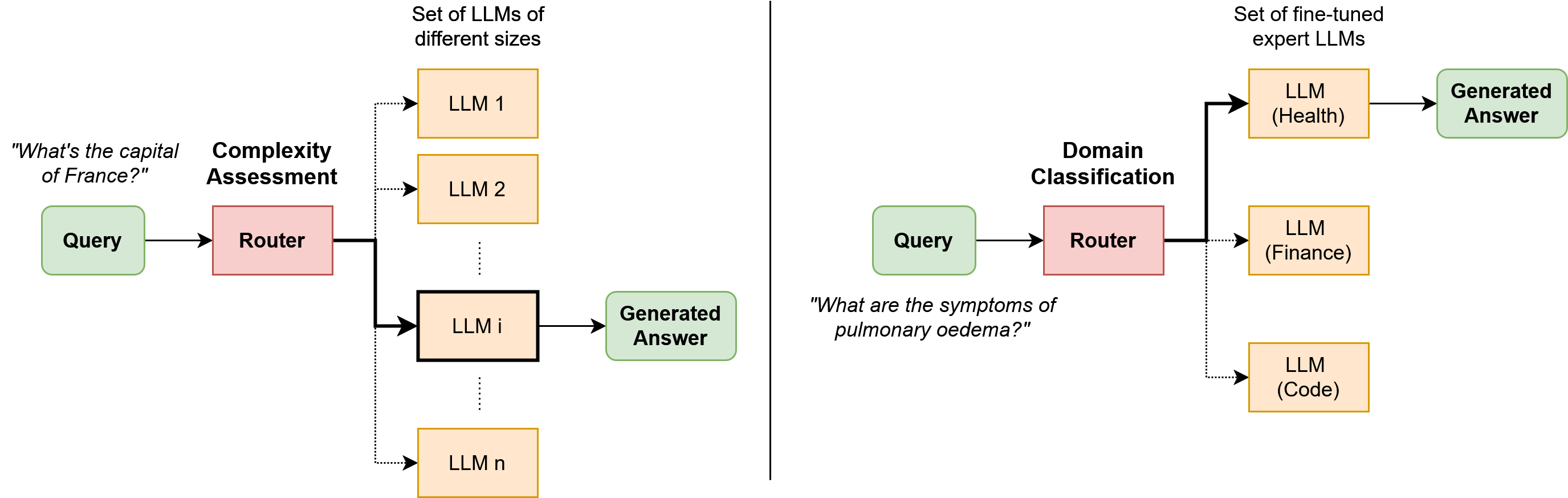}%
  }
  \caption{\textbf{Routing as pre-generation step} -- Before generating an answer, each LLM's ability to provide an appropriate answer is assessed based on the user query complexity and/or topic. \textit{Dotted arrows represent non-selected LLM candidates. Rectangles with straight edges represent the router and routing candidates, while rectangles with rounded corners represent input/output components (user requests and results).}}
  \label{fig:fig1}
\end{figure*}

In \textbf{pre-generation routing}, the system estimates in advance whether an LLM is likely to provide an adequate answer (see Figure \ref{fig:fig1}). This approach minimises latency, as no model needs to generate a response before routing decisions are made. There are two main approaches to achieve this, depending on our use case and LLM setup: 
\begin{itemize}
    \item infer the domain of knowledge of the query and route it to the associated LLMs trained as domain experts
    \item assess the LLM candidate's ability to answer a query of a given complexity and then route the query to the LLM with sufficient reasoning ability.
\end{itemize}

In this survey, we define the complexity of a query as an estimate of the degree of difficulty that an LLM is likely to encounter in the generation of a response to a given query. Within the context of RAG-based conversational agents, complexity can be categorised as follows: 
\begin{itemize}
    \item \textit{low complexity user query} may consist of a simple greeting that does not require retrieval
    \item \textit{intermediate complexity user query} might involve extracting explicit information from a single document
    \item \textit{high complexity user query} may necessitate the extraction of implicit information from multiple documents through reasoning.
\end{itemize}

\subsection{Routing as a Post-Generation Step}\label{Cascading: Routing as a Post-Generation Step}

\begin{figure*}[!htbp]
  \centering
  \makebox[\textwidth][c]{%
    \includegraphics[width=0.8\textwidth]{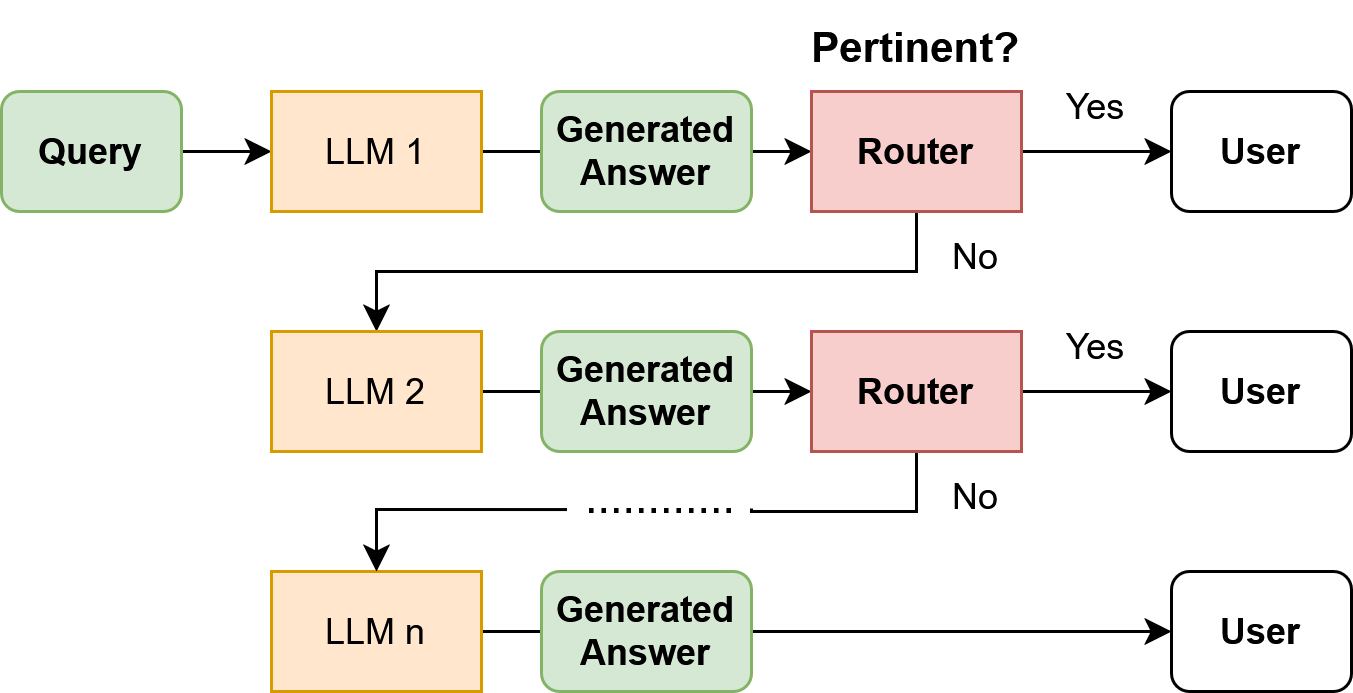}%
  }
  \caption{\textbf{Routing as post-generation step (or cascade routing)} -- The relevance of a larger LLM is determined by the evaluation of the answers generated by the current LLM. Each candidate response is evaluated sequentially. If an answer is deemed inadequate or untrustworthy, the user query is routed to a larger LLM. Typically, the cascade sequence is static. \textit{Rectangles with straight edges represent the router and routing candidates, while rectangles with rounded corners represent input/output components.}}
  \label{fig:fig2}
\end{figure*}

One might conceptualise routing as a \textbf{post-generation step}, where each model response is assessed iteratively (or in a cascade manner) by selecting progressively more advanced models until the response is considered adequate (see Figure \ref{fig:fig2}). Unlike pre-generation routing, cascade routing evaluates generated responses. \cite{kamalloo-etal-2023-evaluating}. This method can incur higher latency and cost, since multiple LLMs may need to generate answers for a single query.\\

A more flexible variant of this approach is the \textbf{multi-choice cascading} method \cite{dekoninck2024}. Although it is still a post-generation technique, it incorporates elements of pre-generation routing by allowing the next model to be selected dynamically. Rather than following a fixed sequence, the router can forward the query to any available LLM based on an evaluation of the current response. While this hybrid strategy generally outperforms simple cascading and traditional pre-generation methods, it still requires multiple generations per query, which impacts cost and latency.

\section{Q3: How routing should be implemented?}\label{how-section}

This section summarises the various strategies for implementing routing in LLM-based systems, as summarised in Table~\ref{tab:meta_category}. We group the methods into four main categories : similarity-based, supervised, reinforcement learning-based, and generative routing. The complexity, resource requirements, generalisation and effectiveness of these strategies differ, offering distinct trade-offs between performance, cost and adaptability.\\

\begin{table*}[!htbp]
\centering
\small
\makebox[\textwidth][c]{%
\begin{tabular}{p{0.25\textwidth}p{0.30\textwidth}p{0.45\textwidth}}
\toprule
\textit{\textbf{Routing Category}} & \textit{\textbf{Routing Strategy}} & \textit{\textbf{Studies}} \\
\midrule
\textbf{Similarity routing} &\textit{ Query similarity} & \cite{jang2023, chen2024routerdc, lee2024, manias2024, malekpour2024, srivatsa2024, stripelis2024}\\
 & \textit{Clustering} & \cite{ pichlmeier2024, srivatsa2024, jitkrittum2025} \\
 & \textit{Preference similarity} & \cite{zhao2024,ong2024, zhang2025} \\
\midrule
\textbf{Supervised routing} & \textit{Recommendation System} & \cite{ong2024,zhang2025, zhuang2025} \\
 & \textit{Domain Classification} & \cite{jain2024,wang2024CoE} \\
 & \textit{Query complexity inf.} & \cite{dekoninck2024, ding2024, hu2024, jeong2024, malekpour2024, ning2024, shnitzer2023, stripelis2024, srivatsa2024, lee2025, ong2024, somerstep2025,wang2025mixllm, zhang2025} \\
  & \textit{Answer confidence inf.} & \cite{chen2023} \\
 & \textit{Knowledge Graph} & \cite{feng2024} \\
\midrule
 \textbf{RL routing}& \textit{Stateless} & \cite{sikeridis2024} \\
 & \textit{State-based} & \cite{nguyen2024, li2025llmbandit} \\
 & \textit{Reward inf.} & \cite{hari2023, lu2024} \\
\midrule
\textbf{Generative routing} & \textit{Prompt} & \cite{shen2023, li2024,ning2024} \\
& \textit{Seq. probability} & \cite{lee2024} \\
& \textit{Token probability} & \cite{ramirez2024} \\
 & \textit{Repeated calls} & \cite{aggarwal2024, yue2024} \\
 & \textit{LLM fine-tuning} & \cite{chai2024,chuang2024, liu2024meswitch, patil2023, ong2024} \\
 & \textit{Code execution} & \cite{zhang2023} \\
\specialrule{1.5pt}{0pt}{0pt} 
\end{tabular}%
}
\caption{Mapping of the different routing categories and strategies associated. \textit{RL: Reinforcement Learning; inf.: inference; Seq.: sequence}}
\label{tab:meta_category}
\end{table*}

\subsection{Similarity-based Routing}
Similarity-based routing routes queries to LLMs based on semantically similar queries from past interactions, using unsupervised or weakly supervised signals. This approach includes techniques based on query similarity, clustering of previous interactions, and human preference similarity. While generally lightweight and adaptable, these methods often face limitations when handling complex queries or generalising across diverse tasks.

\subsubsection{Query similarity}

This is typically achieved by computing the cosine similarity between query embeddings. The fundamental premise is that similar queries should be processed by the same model. The simplest implementations use 1-Nearest-Neighbour routing \cite{stripelis2024}, yet these approaches fail to capture complex query-response relationships and often perform worse than random baselines. Leveraging multiple nearest neighbours address this limitation \cite{manias2024,jang2023}\footnote{\href{https://github.com/aurelio-labs/semantic-router}{aurelio-labs/semantic-router}} by performing similarly to prompt-based intent detection achieving around 90\% accuracy \cite{manias2024}, while \citet{jang2023} demonstrated strong performance on non-generative tasks but inferior results on generative tasks. This disparity in performance across different types of task reveals that the effectiveness of routing based on query similarity is highly dependent on the type of task and may be influenced by the complexity of the query.\\ 
In the domain of text-to-SQL generation, \citet{malekpour2024} proposed selecting the cheapest LLM that exceeds a success threshold, defined as the minimum proportion of similar queries for which the LLM returned the correct SQL query. Their method achieved near-optimal performance of 60.1\%, compared to 61.0\% for the best-performing standalone model (\textit{GPT-4o}), while reducing costs by a factor of 1.1. This modest reduction reflects the predominant routing of queries to \textit{GPT-4o} (81\%), which raises concerns about cost-effectiveness: a routing strategy must avoid achieving higher quality by exclusively routing to the larger LLM. Moreover, the routing strategy did not include a specialist code generation model, which could potentially have offered similar quality at lower cost.\\ 

The performance of similarity-based routing can be further improved by optimising query representation through contrastive learning \cite{lee2024, chen2024routerdc}. This technique ensures that queries with similar meanings are embedded close together in the representation space, while queries with different meanings are placed further apart \cite{LeKhac2020}. While this addresses the semantic representation problem, it introduces substantial training overhead and requires domain-specific tuning. \\ 
In the context of task-oriented dialogue, where systems must extract user intents and track dialogue context, \citet{lee2024} apply contrastive learning within a framework called \textit{OrchestraLLM}. \textit{OrchestraLLM} shows a slight improvement in accuracy compared to a direct call to the larger LLM while reducing the number of calls to this model by 50\% to 80\%. \citet{zhao2024lora} extends contrastive learning-based routing to select the most appropriate LoRA adapter for a given task. With this approach, the authors dynamically switch LLM capabilities instead of deploying multiple LLMs by selecting the most appropriate LoRA adapter. \citet{zhao2024lora} found that their contrastive learning-based LoRA adapter routing strategy was particularly effective for natural language understanding tasks such as translation or description generation from structured data. For natural language processing tasks such as summarisation and sentiment analysis, their approach showed either superior performance or performance comparable to an average pooling of multiple adapters retrieved using their method. \\ \citet{chen2024routerdc} present a more sophisticated approach called \textit{RouterDC}, which applies multi-level contrastive learning to enhance query representation, using multi-level contrastive learning to map both queries and LLMs into shared embedding spaces. Except for the MMLU dataset, \textit{RouterDC} outperforms the best stand-alone LLMs and performs as well as or better than other routing methods, like \textit{Zooter} \cite{lu2024} or supervised approaches, on different datasets. It achieves the highest average performance across out-of-distribution datasets by effectively approximating the results of the best-performing models on each dataset.

\subsubsection{Clustering previous interactions}

The clustering approach treats routing as a two-stage process. First, it identifies the semantic cluster that best represents a query. Then, it selects the LLM that has historically performed best for that cluster type.\\
Multiple implementations using \textit{k-means} clustering have studied this strategy, though with varying degrees of success. The Expert Router framework \cite{pichlmeier2024} leverages pre-trained clusters from large-scale datasets, while other approaches \cite{srivatsa2024, jitkrittum2025} train clusters from scratch. While \citet{pichlmeier2024} do not specify how LLMs are assigned to clusters, \citet{srivatsa2024} and \citet{jitkrittum2025} evaluate which LLM most frequently produces responses matching the ground truth on each cluster.\\ When \citet{srivatsa2024} trained a \textit{k-means} model from scratch for the routing task, the resulting clusters failed to generalize well from training to test datasets. This indicates that both the size and diversity of the training set are critical for effective clustering and generalisation on out-of-distribution tasks. The choice of encoding strategy, whether using a dense representation strategy with \textit{RoBERTa} \cite{zhuang2021} or a sparse strategy with \textit{TF-IDF}, did not significantly affect the results. 

\citet{jitkrittum2025} proposed a generalisable routing strategy which does not require retraining when introducing new LLM candidates. The approach uses \textit{k-means} clustering to form clusters from a dataset. Then, they evaluate LLM performance on the cluster examples.
They calculate the average performance of each LLM across every cluster, representing each LLM as a performance vector amongst clusters. During inference, the system routes new queries to the LLM with the lowest error rate for the nearest cluster to the query. This strategy is dynamic in that the cluster-level performance representation is independent of the original pool of candidate LLMs. Introducing a new model only requires computing its per-cluster error vector on the different clusters. This approach is successful in generalising when different LLMs are used for training and testing; the performance remains comparable to setups involving all candidate LLMs. Furthermore, learning a more sophisticated clustering assignment map does not improve upon the unsupervised approach.

\subsubsection{Preference similarity}

An alternative approach uses human preference data to predict which model users would favour for similar queries. This strategy redefines routing as a problem of estimating probabilities: given the similarity of a query to previous examples, what is the likelihood that a more expensive model would be preferred? Both strategies used preference data to develop ranking-based algorithms.\\ \citet{ong2024} used preference data from the Chatbot Arena Leaderboard \footnote{\href{https://chat.lmsys.org/}{Chatbot Arena Leaderboard}}, a preference-based LLM ranking interface, to propose \textit{RouteLLM}: a range of routing strategies based on user preferences\footnote{\href{https://github.com/lm-sys/RouteLLM}{lm-sys/RouteLLM}} \footnote{\href{https://huggingface.co/routellm}{https://huggingface.co/routellm}}. They reformulate the routing task between a smaller LLM, \textit{Mixtral-8x7B}, versus a larger LLM, \textit{GPT-4-1106-preview}, as predicting the probability that the larger model will be preferred for a given query. To compute this probability, the authors employ a Bradley-Terry (BT) algorithm \cite{bradleyterry1952} with similarity-weighted ranking. They weight the queries from the training dataset according to their similarity to the user query and subsequently use these weighted queries to learn the BT coefficients. These coefficients then allow the estimation of the probability. This strategy is influenced by the embedding strategy as \citet{zhang2025} compared a strategy where they simulate user preferences by exploiting uncertainty through semantic entropy. In their small-vs-large LLM setup, the model with lower semantic entropy is assumed to be preferred. Their enhanced similarity-weighted ranking strategy shows clear improvements over \citet{ong2024}'s original approach, requiring only 27\% of queries to be routed to the larger model on MT-Bench in order to boost the smaller model’s performance by 50\%.

In parallel, \citet{zhao2024} introduced an ELO-based algorithm called \textit{Eagle} \cite{eloranking}. This method incorporates a two-level structure: The global ranking is computed using all available pairwise comparisons to establish an overall model hierarchy. The local ranking is based on pairwise preferences among the nearest neighbor queries, determined via cosine similarity. For any given query, Eagle selects the optimal LLM by computing a weighted sum of its global and local ELO scores, while ensuring the chosen model stays within the user’s budget constraints. Notably, these preference-based methods consistently outperform traditional classification approaches (e.g., SVM, MLP, KNN, BERT-based classifier or LLMs fine-tuned for classification) \cite{zhao2024, ong2024}, suggesting that similarity-weighted preferences capture routing requirements better than standard supervised learning. However, the original RouteLLM strategy performed only slightly better than random routing (56\% vs. 51\%) on one of the experiment of \citet{zhao2024}. The authors suggest that this modest improvement may result from the inherent unreliability of human judgments in preference data \cite{zhang2025}.\\
These results highlight the potential of using information about user preferences to train routing algorithms.

\subsection{Supervised Routing}\label{low-supervised}

Similarity-based routing frequently fails on complex tasks due to its unsupervised nature, particularly when discriminating between similar tasks or when noise levels are substantial. Once LLM performance profiles have been established for different query categories or knowledge domains, routing can be treated as a supervised task.

\subsubsection{Recommendation Systems}

Recommendation systems align with the routing framework philosophy by aiming to select the most suitable LLM for each specific query. It has been studied matrix factorisation, a well-known algorithm in recommendation systems \cite{zhang2022}, which infers latent features from user-item interactions to predict preferences \cite{ong2024, zhuang2025, zhang2025}. \\
This approach extrapolates users preferences between models and queries by learning a hidden scoring function that reflects user preference patterns \cite{ong2024}. Enhanced implementations incorporate uncertainty signals through semantic entropy \cite{zhang2025} or reconstruct comprehensive performance matrices across multiple benchmarks \cite{zhuang2025}.\\
Matrix factorisation proves more efficient than classification-based approaches for cost-quality trade-offs, with implementations achieving 80\% quality gains while requiring only 30\% of calls to \textit{GPT-4-1106-preview} \cite{ong2024}. In contrast, the random assignment required 78\% of calls to this model. However, the effectiveness of this approach varies significantly depending on the implementation and evaluation conditions. Some studies have found that it achieves only minimal improvement over random routing \cite{zhang2025}. \citet{zhang2025} improved performance by incorporating uncertainty signals into the embedding inputs. \citet{zhuang2025} confirmed these findings, demonstrating that this strategy approaches the performance of an Oracle router while scaling to over 100 candidate models.
 
\subsubsection{Domain Classification}

Domain-based routing routes queries to specialized models based on their knowledge domain, such as health, mathematics, coding, etc. This strategy focuses on broad categorical knowledge rather than query-specific details.\\
\textit{BERT}-based implementations, like fine-tuned \textit{DeBERTA-v3-large} model \cite{he2023debertav3i, simonds2024} demonstrate the effectiveness of this approach \cite{simonds2024, wang2024CoE}, with systems like \textit{MoDEM} achieving accuracy  comparable to that of larger, generalist, stand-alone LLMs, while routing to domain-specific expert models \cite{simonds2024}, but at lower costs. \citet{simonds2024}'s approach achieved an MMLU accuracy of 87.7\%, which is closer to that of \textit{Llama-3.1-405B} \cite{grattafiori2024llama3herdmodels} (88.6\%) than\textit{ Qwen 2.5-72B} \cite{yang2024qwen2} (86.1\%), while using a pool of expert 70B models at a cost per token four times lower than that of the 405B model \cite{simonds2024}.\\ \textit{BERT}-based models are not strictly necessary for effective domain-based routing. \citet{jain2024} proposed a two-stage routing strategy called \textit{Composition-of-Experts}. In the first step, a \textit{k-NN} model maps queries to knowledge domains. If the entropy-based measure of classifier uncertainty exceeds a predefined threshold, the model assigns the user query to the 'general' category. In the second step, a mixed integer linear program allocates categories to experts, aiming to minimise costs while adhering to budget constraints. For certain datasets, such as Arena-Hard or MT-Bench, this framework achieved scores comparable to LLMs, like \textit{Llama-3-70B-Instruct} \cite{grattafiori2024llama3herdmodels} and \textit{Qwen2-72B-Instruct} \cite{yang2024qwen2}, using fewer average active parameters (approximately 30-50 compared to 70).\\ However, domain-based routing has significant limitations when queries do not fit into predefined categories. Performance often deteriorates substantially in 'other' or general domains, dropping from 77–97\% accuracy in specialised areas to as low as 53\% for 'other' queries \cite{simonds2024}. This limitation highlights the importance of a thorough domain definition and the difficulty of dealing with edge cases in categorical approaches.\\

Focusing specifically on the knowledge domain rather than the individual complexity of queries improves the router’s ability to generalise on out-of-distribution datasets \cite{wang2024CoE}. This approach allows less restrictive routing rules by linking a routing candidate to more general concepts (i.e., the knowledge domain), resulting in a 3.6\% increase in accuracy compared to the direct mapping of queries to experts.

\subsubsection{Query complexity inference}

Moving beyond domain classification, complexity-based routing recognises that the intrinsic difficulty of individual queries often matters more than the topical domain for routing decisions, and attempts to capture this. By categorising user queries into complexity levels, the routing process becomes a classification task \cite{wang2024, ding2024, malekpour2024, stripelis2024, srivatsa2024}. Query-level routing significantly outperforms domain-based approaches on related datasets, achieving 64.3\% compared to 52.2\% accuracy, and even surpassing much larger individual models \cite{wang2024}. Remarkably, despite using a set of small LLMs (under 9B parameters), it surpasses the performance of a much larger LLM, \textit{Llama-3-70b} \cite{grattafiori2024llama3herdmodels}. However, this improvement comes at the cost of generalization ability, with reduced performance on out-of-distribution datasets. 

The definition of query complexity represents a critical design choice that fundamentally shapes system performance. The first approach measures complexity through performance differences between models of varying sizes\citet{ding2024} by computing the \textit{BART} score \cite{yuan2021}. This score calculates the probability of generating a sequence given a reference sequence with a seq2seq model, called \textit{BART} \cite{lewis2020}. When parameter size differences are substantial, such as between \textit{FLAN-T5} (800m) \cite{chungflant5} and \textit{LLaMA-2-13B} \cite{touvron2023}, 40\% of queries were successfully routed to the smaller model with only 10\% quality degradation. More remarkably, when differences are subtle, such as between \textit{LLaMA-2-13B} and\textit{ GPT-3.5-turbo}, 20\% of queries could still be handled by the smaller model with less than 1\% quality reduction.\\
Alternative complexity definitions focus on previous successes rather than sequence probability \cite{malekpour2024, stripelis2024}. \citet{malekpour2024} demonstrated that classifying the cheapest LLM capable of handling a query could reduce costs by 1.4 times, albeit with a loss of nearly six performance points compared to \textit{GPT-4o-mini} (55.2\% vs 61.0\%). They showed that this strategy resulted in lower successful SQL generation than the similarity retrieval method discussed previously. However, it is important to note that this classification strategy relied more on smaller LLMs, which performed poorly in generating SQL queries. A better selection of LLM might yield different results (i.e., SQL Expert LLM). \citet{stripelis2024} reduces costs by 30\% and latency by 40\% compared to stand-alone generalist LLM (\textit{Fox-1.6B} \cite{hu2024fox1}) or experts LLM across various domains. \\
The robustness-based approach to complexity assessment proposed by \citet{srivatsa2024} evaluates LLM adequacy by generating multiple responses and comparing them to reference answers.\footnote{\href{https://github.com/kvadityasrivatsa/llm-routing}{kvadityasrivatsa/llm-routing}} An LLM is considered adequate for a query if the majority of its responses align with ground truth. Although the router exhibited reduced latency, it did not achieve the same accuracy as the best-performing LLM, \textit{gemma-7b}, suggesting that robustness may not fully capture query complexity nuances.\\

Complexity inference is treated as an optimisation problem in more sophisticated approaches. \citet{wang2025mixllm} proposed \emph{MixLLM}, a method to infer a cost-quality score and route to the routing options that maximise this score. It infers cost-quality scores using fine-tuned embeddings that minimize distance between sentences from the same knowledge domain. Then, they infer the output length for each query and each LLM and calculate the expected cost. Their weighted linear combination of quality, cost, and uncertainty scores, adjusted for user budget constraints, achieved 97\% of GPT-4 accuracy at only 24\% of the associated cost. It outperformed several methods proposed in the survey, including \textit{RouterBench} \cite{hu2024}, \textit{Zooter multi-perceptron} \cite{lu2024}, \citet{sakota_2024}'s supervised strategy, BERT-based multi-classifier \cite{ong2024}, \textit{MetaLLM} \cite{nguyen2024}, \textit{OptLLM} \cite{liu2024optllm}, and \textit{AutoMix} \cite{aggarwal2024}.

Direct performance score inference represents another approach, treating routing as a regression rather than classification task \cite{hu2024}. \citet{somerstep2025} proposed to infer the expected cost and quality of LLM for specific queries using either a \textit{KNN} or a fine-tuned \textit{RoBERTa} model.
Their strategy achieved similar accuracy (95\%) to \textit{GPT-4} on RouterBench \cite{hu2024} but at 20\% of its cost. Both the \textit{KNN} and fine-tuned \textit{RoBERTa} \cite{zhuang2021} models showed comparable performance, each outperforming Ong's preference-based matrix factorisation and fine-tuned \textit{RoBERTa} multi-classifier \cite{ong2024}. \citet{hu2024} derived performance scores as $Performance_{i,j} = \lambda \cdot P_{i,j} - cost_{j}$, where $P_{i,j}$ represents the predicted ability of model $M_j$ to answer query $q_i$, $\lambda$ represents user willingness to pay, and $cost_j$ denotes the model's cost \footnote{\href{https://github.com/withmartian/routerbench}{withmartian/routerbench}}. This regression-based routing achieved performance comparable to \textit{GPT-4} or \textit{Claude V2} on standard datasets while significantly reducing costs. \citet{sakota_2024} extended this concept by proposing three selection strategies: maximising performance regardless of cost, selecting models that exceed performance thresholds while minimising expense, or optimising within budget constraints through integer linear programming. Maximising the performance score without considering cost achieves an accuracy equivalent to the best-performing model, \textit{text-davinci-2}. However, this approach incurs a cost comparable to routing to the highest-performing model, resulting in an 11\% cost reduction. By implementing either the threshold approach or the cost-sensitive method, they achieved comparable accuracy at a significantly reduced cost (approximately 62\% decrease). \citet{liu2024optllm} introduced \emph{OptLLM}, a routing strategy that first infers the expected accuracy of each candidate LLM for a given query. Then, it constructs a set of Pareto optimal solutions using heuristic-based multi-objective optimisation, simultaneously maximising performance and minimising cost. During inference, the query can be routed to the Pareto-optimal solution that meets the user specific requirements, such as budget constraints. Across different datasets, \textit{OptLLM} achieves accuracy comparable to the best single LLM while significantly reducing computational cost.\\
The challenge of generalising to tasks that have not been seen before has led to the development of approaches for out-of-distribution routing. \citet{shnitzer2023} implemented a collection of binary classifiers to determine whether a model $M_i$ within a set of $n$ LLMs $\mathcal{M} = \{M_1,...,M_n\}$ could provide an answer that matches the reference answer for a given query. However, the methodology for labelling the training set is not explicitly described. This model estimates the probability that a binary classifier's prediction is correct for a specific data point within a given task, reflecting the model's uncertainty. They demonstrated that these strategies outperformed the best model on average, \textit{Llama-2-70B} \cite{touvron2023}, by enabling the selection of smaller models that could provide adequate answers.\\

The architecture used for supervised routing varies significantly. When explicitly detailed, most studies employ \textit{BERT}-family models as backbones, adapting them for supervised training: \textit{BERT-based} \cite{devlin2019, ong2024, stripelis2024}, \textit{RoBERTa} \cite{zhuang2021, srivatsa2024, ning2024}, \textit{DistilBERT} \cite{sanh2020distilbert, srivatsa2024, sakota_2024, malekpour2024}. The choice of backbone model highly influence the effectiveness of these classification architectures: \citet{feng2024} demonstrated that replacing the \textit{DeBERTa} model \cite{he2023} with the larger \textit{RoBERTa} architecture \cite{zhuang2021} enhanced the performance of the \textit{HybridLLM} framework. Alternative implementations have employed \textit{T5} \cite{raffel2020,srivatsa2024, jeong2024}, \textit{multi-layers perceptron} \cite{hu2024, stripelis2024, lee2025, zhang2025, somerstep2025}, \textit{Random Forest} \cite{srivatsa2024}, \textit{optimisation programmes} \cite{dekoninck2024, wang2025mixllm, liu2024optllm}, or \textit{K-NN} \cite{shnitzer2023,stripelis2024, hu2024, zhang2025, somerstep2025}. One notable difference from other BERT-based approaches is \citet{mohammadshahi2024}'s \textit{Routoo} Orchestrator\footnote{\href{https://github.com/Leeroo-AI/leeroo_orchestrator}{Leeroo-AI/leeroo\_orchestrator}}, which employs decoder-only LLMs for query encoding rather than bidirectional language models (i.e., BERT models \cite{devlin2019}). This approach leverages the autoregressive nature of these models by extracting representations from predefined last tokens (typically '<\textbackslash s>' or '<EOS>'). These embedding strategies 
have demonstrated high performance, with top-ranked models on the MTEB leaderboard\footnote{\href{https://huggingface.co/spaces/mteb/leaderboard}{MTEB leaderboard}} being decoder-only embedding models, such as Nvidia's \textit{NV-Embed-v2} \cite{lee2024nvembed} and BAAI's \textit{bge-en-icl }\cite{li2024makingtextembeddersfewshot}. However, these require significant computing resources due to their larger size. Decoder-only models are typically fine-tuned for dense retrieval \cite{ma2024, wang2024improving-text, li2024llama2vec, li2024makingtextembeddersfewshot, lee2024nvembed, muennighoff2024}. The researchers improve model selection by identifying a subset of LLMs that perform best across a given set of queries, aiming to select models whose strengths complement each other for more effective routing. The Routoo approach demonstrated superior accuracy on MMLU \cite{hendrycks2021measuring} compared to larger standalone LLMs like \textit{Llama2-70B }(75.9\% vs 69.9\%) and \textit{Mixtral-8x7B} (75.9\% vs 70.6\%) at similar or lower costs. Although this strategy employs lightweight methods for query categorisation, it requires significant and unnecessary resources since it necessitates hosting and running a large decoder-only model for encoding.\\

Supervised pre-generation routing success depends on complementary capabilities among available LLMs \cite{srivatsa2024}. This limitation is discussed in the Section \ref{Using complementary routing options}. \citet{yue2024} found that a \textit{RoBERTa} model fine-tuned to route between \textit{GPT-3.5} and \textit{GPT-4} was less effective than prompt-based routing, indicating that routing candidate selection may influence supervised routing effectiveness. The supervised routing paradigm extends beyond simple generation tasks to encompass guardrails \cite{lee2025}, retrieval strategies \cite{jeong2024}, and prompting techniques \cite{ning2024}. \citet{lee2025} implemented \textit{SafeRoute}, dynamically switching between guardrail models based on complexity, achieving the larger model's F1 score while significantly reducing latency. \citet{jeong2024} demonstrated similar results with their \textit{Adaptive-RAG}\footnote{\href{https://github.com/starsuzi/Adaptive-RAG}{starsuzi/Adaptive-RAG}} framework, routing between no-retrieval, single-step naive RAG, and multi-step RAG methods based on complexity assessment. Similarly, in \citet{ning2024}'s study, the model determines whether a query can be answered using complex prompting approach, treating it as a binary classification task\footnote{\href{https://github.com/imagination-research/sot}{imagination-research/sot}}. The authors found that incorporating a router between prompts is more effective than applying their complex prompting to all queries. However, no comparison was made between the efficacy of their approach and that of a basic prompt as a baseline. Therefore, it is difficult to ascertain the relative efficacy of this approach.

 \subsubsection{Knowledge Graphs}
 
The main drawback of most of the routing methods discussed is that they cannot be generalised to new routing options.
As noted by \citet{feng2024}, these methods rely on a transductive learning framework, which involves learning specific rules from a corpus and applying them to particular cases, similar to those encountered during training. In rapidly evolving environments where new options frequently emerge, maintaining such architectures becomes prohibitively costly, necessitating frequent retraining whenever new models are introduced.

To address this generalisation challenge, \citet{feng2024} proposed an inductive graph framework, \textit{GraphRouter}. This framework focuses on learning contextual information regarding task interactions and models, ultimately enhancing generalisation. The graph consists of three types of nodes: task, query, and LLM. To initialise task and LLM nodes, they first generate their descriptions and include additional calling cost information for the LLM nodes. Then, they encode these descriptions using a \textit{BERT}-like model \cite{devlin2019}. Query nodes are also encoded using the same model. Edge features capture relationships between nodes. Task-query edges indicate the relevance of queries to specific tasks (e.g. maths-related queries in GSM8K contexts), while LLM-query edges provide performance scores adjusted for desired costs by combining cost and performance metrics.
The system implements a two-layer graph attention network with 32-dimensional hidden layers to update node representations using local neighborhood information. Subsequently, selecting an LLM can be reframed as an edge prediction between the query and LLM nodes.\\ They demonstrated a performance surpassing largest standalone LLM on a multi-task dataset while requiring lower costs. Their results also exceeded those of prompt-based routing, \textit{HybridLLM} \cite{ding2024}, \textit{FrugalGPT} \cite{chen2023}, and a bandit-based model. Notably, with new LLM options, they achieved a performance improvement of 4 points over \textit{FrugalGPT} and 21 points over \textit{HybridLLM}.

\subsubsection{Answer confidence inference}

Cascading approaches represent a fundamentally different routing paradigm, in which routing decisions are made sequentially rather than through pre-generation classification. In this framework, models at each iteration estimate the likelihood that generated answers are correct, with routing decisions based on confidence thresholds rather than query characteristics. \citet{chen2023} proposed \textit{FrugalGPT}\footnote{\href{https://github.com/stanford-futuredata/Frugalgpt}{stanford-futuredata/Frugalgpt}}, which infers the probability that the generated answer is correct using a \textit{DistilBERT} regression model \cite{sanh2020distilbert}. The authors reported that their framework could save between 59\% and 98\% of costs while maintaining similar accuracy to larger standalone models, such as \textit{GPT-4}. This strategy is often used for comparisons but underperforms against alternatives, such as LLM-based repeated calls routing \cite{aggarwal2024} and graph-based supervised routing \cite{feng2024}. Moreover, even lower-resource LLM-based strategies, such as the token probability method \cite{ramirez2024}, match its efficacy.

\subsection{Reinforcement Learning-based Routing}\label{low-reinforcement}
Supervised or similarity-based may not be able to adapt to new routing options or a new context, such as changes in the way users express themselves. This limitation arises because the routing process needs to learn from interactions \cite{Sutton2014}. Reinforcement learning (RL) addresses this issue by treating routing as a sequential decision-making problem, whereby the router learns to select the optimal model through environmental feedback \cite{Sutton2014}. In this paradigm, routing is framed as the selection of actions (models) based on the current state (user queries), with the aim of maximising rewards while respecting cost constraints.

\subsubsection{Stateless algorithms}

Stateless algorithms offer a low-resource alternative by directly mapping rewards to actions. The Stochastic Learning Automaton (SLA) of the \textit{PickLLM} framework demonstrates how this strategy works by maintaining probability distributions over candidate models and updating them based on interaction outcomes \cite{sikeridis2024} using a basic linear reward-inaction scheme \cite{narendra1974}: increasing the probability of selecting the LLM with the highest reward while decreasing the probability of other candidates. This strategy significantly reduces cost and latency compared to using the most expensive option or randomly selecting a model. Specifically, the SLA option reduces latency more consistently than the Q-learning option. Regarding accuracy, the model performed slightly better than \emph{Llama-2-70B} and \emph{Mixtral-8x7B} \cite{jiang2024mixtralexperts} on a medical subset and a computer science subset. However, especially when compared to \emph{Mixtral-8x7B}, it underperformed on a subset of Reddit. This discrepancy may be due to the absence of \emph{Mixtral-8x7B} in the provided set of options.

Instead of relying on action probabilities, one could optimise selection using expected rewards, referred to as \emph{Q-values} \cite{kaelbling1996}. In Q-learning, an agent iteratively updates its \emph{Q-values} based on the rewards received by interacting with the environment \cite{kaelbling1996}. The agent selects the LLM candidate with the highest expected reward, though it suffers from exploitation bias where successful actions dominate future selections Epsilon-greedy exploration mitigates this by balancing known good choices with random exploration \citet{sikeridis2024}. This technique selects a random action with a probability of $\epsilon$ and chooses the action with the highest expected reward, as indicated by \emph{Q-values}, with a probability of $(1 - \epsilon)$ \cite{sikeridis2024}. This strategy effectively reduces both cost and latency compared to purely random selection or routing only to standalone LLMs. Performance remains comparable to SLA approach in terms of accuracy.

Stateless strategies effectively converge on the LLM candidate that best fits the data, resource requirements, or context at a given time $t$ \cite{sikeridis2024}. However, they are poorly suited for real-world production environments where context evolves over time. These methods cannot adapt their actions to specific contextual changes

\subsubsection{State-based algorithms}

State-based approaches are a significant conceptual advance because they integrate contextual information into routing decisions. Rather than learning a single optimal model, these systems determine which models perform best in specific situations. \\ \citet{nguyen2024} proposed the \textit{Meta-LLM} framework, which incorporates a contextual multi-armed bandit (MAB) model with a discrete set of actions, or arms, that can be selected at each step. Each arm represents a candidate LLM that the router may choose. Since the reward associated with each arm is initially unknown, the model learns over time by selecting an arm, observing the resulting rewards, and updating its expected reward. The goal is to learn to select the optimal arm in order to maximize cumulative reward over time. The MAB model achieved comparable accuracy to the most expensive and high-performing LLM, \textit{text-davinci-002} (91.0 versus 90.9), while incurring significantly lower costs (0.3 versus 4.8 \$ per 10,000 queries). The authors explain that some queries were correctly classified by small LLMs but not by large ones. In another trials, MAB matched the performance of the most performant LLM, \textit{Llama-2-7B} (91.6 versus 91.5), and even outperformed the most expensive model, \textit{Claude Instant} (91.6 versus 87.6), while requiring drastically lower costs: \textit{MetaLLM} at 0.5 \$ per 10,000 queries, \textit{Claude Instant} at 2.5 \$ per 10,000 queries, and \textit{Llama-2-7B} at 2.4 \$ per 10,000 queries. 

\citet{li2025llmbandit} extended contextual bandits by creating model identity vectors that encode performance patterns across standardised examples, in a manner similar to the approach adopted by \citet{jitkrittum2025}. The primary goal is to learn a performance representation based on a consistent corpus to obtain future candidate representations by running them on the same sample. The algorithm incorporates user preferences through proximal policy optimisation \cite{schulman2017}. When comparing this strategy with an implementation of \textit{RouteLLM} \cite{ong2024}, their approach demonstrated superior performance. Notably, the \textit{RouteLLM} implementation did not outperform random routing.
Future research should compare the complexity of this reinforcement learning-based architecture with the simplicity of the \citet{jitkrittum2025} approach to determine whether such sophisticated reinforcement learning techniques are necessary to achieve dynamic routing.

\subsubsection{Reward-based inference}

\citet{hari2023} noted that creating a Q-table to map the actual reward $r_{M_i}$ for selecting an LLM $M_i$ over all queries would be impractical. Instead of learning through direct interaction, these methods use separate reward models to predict which LLM will perform best for a given query.
In their \textit{Tryage} framework, \citet{hari2023} infer expert model losses (rewards) by training a \textit{BERT-small} model \cite{turc2019, bhargava-etal-2021-generalization} on a dataset containing queries and corresponding expert model losses. This strategy was taken a step further by \citet{lu2024} with \textit{Zooter}, a regression model trained from \textit{mdberta-v3-base} \cite{he2023} through knowledge distillation from a reward model, \textit{QwenRM} \cite{bai2023qwentechnicalreport}. Using a reward model proves to be an effective approach to the routing problem. \citet{hari2023} discovered that their framework achieved greater routing effectiveness to the ``ideal'' expert model (50.8\% accuracy) compared to a stand-alone \textit{GPT-3.5} (23.6\%) or the fine-tuned LLM \textit{Gorilla} \cite{patil2023} (10.8\%). There are no explicit details on the selection criteria for the ideal expert model.

Beyond demonstrating technical feasibility, \citet{lu2024} revealed two critical insights about the design of effective routing systems. Firstly, the complementarity of models matters. The scores of the independent models vary significantly depending on the dataset, with some models excelling on specific datasets; for instance, \textit{WizardLM} performs well on the Flask Dataset \cite{ye2024flask}, while \textit{Llama-2-Chat} \cite{touvron2023} performs on AlpacaEval.
Nevertheless, Zooter achieves an average performance comparable to the best model across each dataset. Secondly, diversity in parameter size becomes crucial for complex tasks, as evidenced by performance gaps when all candidate models were of a similar size. In a benchmark comprising MMLU, GSM8K, and HumanEval, all open-source models exhibited poor performance, which constrained \textit{Zooter}'s overall effectiveness. However, \textit{Zooter}'s performance closely aligns with that of \textit{GPT-4}, apart from the last benchmark, where \textit{GPT-4} significantly outperforms it (32.3 for \textit{Zooter} versus 88.3 for \textit{GPT-4}). Changing the reward model ranking did not improve \textit{Zooter} performance on this particular benchmark. This difference may be related to the tasks' complexity, which may be too challenging for 13B parameter models, given that \textit{GPT-4} is much larger than the LLMs integrated into the router.

\subsection{Generative-based routing}\label{low-llm}
This category of strategy uses the emerging capabilities of LLMs to make routing decisions, taking advantage of their ability to perform unsupervised multitasking and generalise to new situations \cite{radford2019language, wei2022emergent}.
 
\subsubsection{Token probability}

The most resource-efficient approach in this category involves measuring output uncertainty at the token level \cite{ramirez2024}. Actually, it achieves this in a low-resource manner by measuring \textit{output uncertainty} or \textit{margin}. Using the list of possible tokens returned by an LLM for the token in the first position of the generation, researchers calculate the margin between the probabilities of the first and second most likely tokens. This method represents a significant improvement in resource requirements over sequence-level approaches like the \textit{BART} score, as it requires only partial generation rather than complete output generation. A larger LLM is called if the margin exceeds a threshold established by a budget-based criterion. This approach has proven to be effective in maximising performance while minimising costs while routing between smaller and larger LLM, including regression models, \textit{HybridLLM} \cite{ding2024} and \textit{FrugalGPT} \cite{chen2023}. The differences are most significant when using the small-large LLM pair \textit{GPT-3} and \textit{GPT-4}. This suggests that the token probability approach is most valuable when there is a significant capability gap between models, allowing clear signals of uncertainty to emerge. \textit{Frugal-GPT} performs better on several datasets of classification tasks (e.g. ISEAR, RT-Pol) with smaller LLM pairs (\textit{Mistral-7B-Instruct-v0.2 \& Mixtral-8x7B-Instruct-v0.1}, \textit{Llama-2-13B}-hf \& \textit{Llama-2-70b-hf}) but not on Q/A and reasoning tasks.

\subsubsection{Sequence probability}

Sequence-level probability assessment represents a natural extension of token-level approaches, considering the cumulative uncertainty across entire generated sequences. \citet{lee2024} proposed using the normalised sequence-level probability of a smaller LLM when determining whether to route a query to a larger LLM. However, their findings reveal a critical limitation of this approach: it tends to rely too heavily on larger models, essentially negating the efficiency benefits that routing is designed to provide. Consequently, this approach is less effective in reducing call frequency to the larger LLM than the supervised methods discussed earlier.

\subsubsection{Prompt-based routing}

Prompt-based routing is the most accessible form of generative routing. It requires minimal infrastructure and leverages the natural language understanding capabilities of LLMs. This approach operates through two distinct paradigms: pre-generation routing via function-calling mechanisms and post-generation routing via verbalised confidence assessment.

In pre-generation scenarios, LLMs are given routing options and task descriptions, allowing them to decide which model to route to based on the characteristics of the query. Routing is performed based on the option returned by the LLM \cite{ning2024, shen2023}. Although it leverages an LLM, it is more energy-efficient and requires fewer resources than fine-tuning an LLM. \citet{shen2023} proposed \textit{HuggingGPT}\footnote{\href{https://github.com/microsoft/JARVIS}{microsoft/JARVIS}}, a framework for task planning and execution \cite{shen2023}. An LLM is prompted to select between different models based on their descriptions. This approach offers the advantage of performing inference with minimal examples in the prompt, known as ``few-shot inference'', or without any examples at all, known as ``zero-shot inference''. This is particularly useful when limited resources are available for annotation.
Studies by \citet{ning2024} demonstrate that even highly performant LLM, such as \textit{GPT-4} did not outperform smaller, task-specific models like fine-tuned \textit{RoBERTa}. This limitation stems from the fact that it is difficult to make accurate routing decisions without observing the model's actual performance in response to a specific query.

This limitation is addressed by post-generation, prompt-based routing, which allows models to express uncertainty after attempting to answer queries \cite{li2024}. This is known as ``verbalised confidence'' \cite{xiong2024}. \citet{li2024} proposed \textit{Self-Route}. They instructed the model to indicate whether additional context, retrieved using a RAG strategy, is sufficient. Otherwise, they suggest using the entire document from which the chunks were extracted. 
Their findings reveal that implementing this strategy outperformed a naive RAG system in terms of accuracy for the most commonly used LLMs (i.e. \textit{GPT-4o} and \textit{GPT-3.5-turbo}). The strategy proposed also outperformed sending large contexts for \textit{GPT-3.5}, which has a smaller context window (16k). In contrast, performance between \textit{Self-Route} and sending a large context proved comparable for\textit{GPT-4o}, which has a larger context window (128k) while using an average of 61\% fewer tokens. Finally, for \textit{gemini-1.5-pro}, which features a context window of 1M tokens, sending only large contexts appears to perform better than \textit{Self-Route}, although the latter's performance remains reasonably close. The results for \textit{gemini-1.5-pro} stem from its large context window, which allows the transmission of extensive contexts without truncation.
While this method shows promise, it suffers from a critical weakness: LLMs tend to be overconfident in their certainty assessments \cite{xiong2024}. This overconfidence can lead to suboptimal routing decisions, with some studies showing that verbalized confidence performs no better than random routing \cite{chuang2024}.

\subsubsection{LLM fine-tuning}

When computational resources permit, fine-tuning LLMs specifically for routing tasks offers the potential for superior performance through task-specific optimization. This approach encompasses several strategies: domain classification\cite{liu2024meswitch}, complexity scoring \cite{ong2024}, and API call generation \cite{patil2023}.

The domain classification approach demonstrates remarkable potential, with \citet{liu2024meswitch} achieving accuracy improvements from 15\% to nearly 100\% on MMLU through supervised fine-tuning of a \textit{Qwen1.5-1.8B-Chat} model \cite{bai2023qwentechnicalreport}.\footnote{\href{https://github.com/godcherry/ExpertTokenRouting}{godcherry/ExpertTokenRouting}} This meta-model categorises the prompt, and the corresponding pre-trained expert associated with that category then generates a response. However, this approach requires retraining when new domains are introduced, limiting its adaptability. In contrast, \citet{ong2024} fine-tuned a \textit{Llama-3-8B} \cite{grattafiori2024llama3herdmodels} on a scoring task to evaluate both the complexity of a query and the model's ability to answer it through LLM evaluation \cite{ong2024} and found that fine-tuning for complexity scoring provided only a small improvement on non-LLM techniques, suggesting that the benefits of fine-tuning depend heavily on the specific task. The authors' differing levels of optimism regarding the fine-tuning of an LLM for classification or regression tasks may also stem from \citet{ong2024} comparing this technique with other optimised approaches, while \citet{liu2024meswitch} evaluated it against the same standalone LLM without fine-tuning. The API call generation approach by \citet{patil2023} treated routing as a code generation problem.\footnote{\href{https://github.com/ShishirPatil/gorilla}{ShishirPatil/gorilla}}. Their fine-tuned\textit{ Llama-7B} model achieved substantial improvements over larger models in zero-shot scenarios, demonstrating that task-specific fine-tuning can enable smaller models to outperform larger general-purpose models in specialized routing contexts. It outperforms larger models after fine-tuning, achieving an average improvement of 35 points over \textit{GPT-3.5-turbo-0301} and 46 points over \textit{GPT-4-0314}.
\\
\subsubsection{Repeated calls}

The repeated calls approach relies on model confidence, which is demonstrated by consistent outputs across multiple generation attempts. By generating multiple responses at elevated temperatures and analyzing their consistency, this method provides robust confidence estimates without requiring additional model training or complex probability calculations. Then, routing to the next model can be triggered when the LLM exhibits low confidence. Smaller LLMs tend to answer simple questions consistently but show inconsistencies when confronted with more complex questions \cite{yue2024}.

Two approaches implemented this approach \cite{aggarwal2024, yue2024}. \citet{aggarwal2024} introduced a three-step approach called \textit{Automix}\footnote{\href{https://github.com/automix-llm/automix}{automix-llm/automix}}. 
First, a smaller model is used to generate an answer to a query based on related context. Next, the same model is prompted with a few-shot meta-prompt, called the verification prompt, to verify whether the answer is consistent with the context. To estimate the model’s alignment with the context, a confidence score is calculated by generating the answer multiple times at a high temperature and determining the proportion of consistent answers. Based on this confidence score, the router either retains the current answer or calls a larger model. This procedure is repeated until the confidence score reaches an acceptable level or all models in the series have been tested. The authors employed two routing strategies: a more complex strategy based on a partially observable Markov decision process (POMDP), and a simpler strategy based on a confidence-cost/quality trade-off threshold. The authors reported that \textit{Automix} \cite{aggarwal2024} achieves a higher F1 score on the QASPER and COQA datasets and superior accuracy across a range of costs compared to \textit{HybridLLM} \cite{ding2024}, \textit{FrugalGPT}, \cite{chen2023} or stand-alone models such as \textit{GPT-4} and \textit{Llama-2-13B} \cite{touvron2023}, particularly the routing based on the POMDP approach. Even in low-resource scenarios with a small training dataset size, their method significantly outperforms both \textit{HybridLLM} \cite{ding2024} and \textit{FrugalGPT} \cite{chen2023}. However, these results require cautious interpretation, as \citet{wang2025mixllm} found that the method did not outperform random routing on RouterBench \cite{hu2024}.\\
The \textit{Mixture-of-Thoughts} approach \cite{yue2024} takes a different perspective, focusing on consistency across different reasoning representations rather than answer correctness \footnote{\href{https://github.com/MurongYue/LLM_MoT_cascade}{MurongYue/LLM\_MoT\_cascade}}. They assessed response consistency across different reasoning-based prompting.\\
In both studies, users have to call an LLM multiple times for each query, and despite their efficiency, these approaches prove to be resource-intensive.

\subsubsection{Code execution}

When it comes to code generation tasks, successful code execution provides an unambiguous confidence signal, eliminating the ambiguities inherent in other assessment methods. \textit{EcoAssistant}\footnote{\href{https://github.com/JieyuZ2/EcoAssistant}{JieyuZ2/EcoAssistant}}, an iterative multi-agent code generator designed to query external knowledge for question and answering, demonstrates this principle by using execution success as a routing criterion, forwarding failed code generation attempts from smaller to larger models \citet{zhang2023}. The authors reported that their framework generates more successful code snippets at a lower cost than using \textit{GPT-4}. Although more expensive than using \textit{GPT-3.5-turbo}, \textit{EcoAssistant} significantly exceeded the percentage of successful code generation.

\section{Discussion}\label{Discussion}
In this work, we described a wide range of routing strategies, from learning similarities to fine-tuning LLM. Most of the strategies presented in this paper follow the pre-generation approach, employing frugal mechanisms such as similarity learning or supervised learning. In contrast, post-generation strategies are generally more resource-intensive, as they often involve generating multiple responses for a single query.
 
This survey demonstrates that the routing problem can be effectively addressed through low-resource solutions that maintain generalisation capabilities without incurring significant financial or computational costs \cite{feng2024}.

\subsection{Industrial Consideration}\label{Industrial Consideration}
In addition to providing a summary of the current state of scientific contributions, it is essential to also consider the industrial landscape. In other words, \emph{what current strategies are companies implementing and disseminating in light of the findings from this review?} 
It seems that most companies aim to direct users queries to specific tools, prompts or scripts. These routing methods are generally based on a rather simplistic methodology, including conditional approach\footnote{\href{https://docs.haystack.deepset.ai/docs/conditionalrouter}{Haystack's ConditionalRouter}, \href{https://docs.haystack.deepset.ai/docs/filetyperouter}{Haystack's FileTypeRouter}}, similarity routing\footnote{\href{https://github.com/aurelio-labs/semantic-router}{aurelio-labs/semantic-router}}, or simply a prompt-based classification system\footnote{\href{https://python.langchain.com/v0.1/docs/use_cases/query_analysis/techniques/routing/}{LangChain's router}, \href{https://docs.llamaindex.ai/en/stable/module_guides/querying/router/}{Llamaindex's router}}
Some companies leverage more sophisticated architecture by employing an LLM to generate synthetic data used to implement a classification-based routing with a small classifier\footnote{\href{https://github.com/lamini-ai/llm-routing-agent}{lamini-ai/llm-routing-agent}}. Many of the mainstream LLM providers propose prompt-based routing, such as AWS\footnote{\href{https://github.com/awslabs/multi-agent-orchestrator}{awslabs/multi-agent-orchestrator}} or OpenAI\footnote{\href{ https://github.com/openai/swarm}{openai/swarm}}. This survey aims to improve the transfer of various lightweight routing strategies identified from research to industry.

\subsection{Key challenges}\label{Key challenges}

\subsubsection{Going beyond financial costs}\label{Going beyond financial costs}

Most studies focus on optimising the trade-off between financial costs and answer quality \cite{chen2023} while overlooking other significant expenses, such as computational and ecological costs. Computational requirements and environmental impacts are intricately linked. Models with larger parameter numbers generally require more floating-point operations, resulting in increased energy consumption \cite{Kaack2022, Desislavov2023}. It is essential to minimise not only the financial costs but also the computational requirements, given the urgent need to mitigate the impact of LLMs on climate change \cite{Kaack2022, Luccioni2024}. This can be accomplished by employing routing modules that activate more resource-intensive components only when necessary. Future work should consider incorporating the computational and environmental costs discussed in section \ref{cost-section} into the cost function $C_{M}$ from the Equation \ref{eq1}. 

\subsubsection{Standardisation of routing strategy experiments}\label{Standardisation of routing strategy experiments}

The field lacks a standardised framework for evaluating routing strategies. Many proposals are only benchmarked against non-routing or self-defined baselines. This makes it difficult to assess or compare their effectiveness objectively. More recently, datasets have been introduced to allow the comparison of different routing architectures under the same conditions: \emph{RouterBench} \cite{hu2024}, \emph{MixInstruct} \cite{jiang2023}, \emph{EmbedLLM} \cite{zhuang2025}, \emph{SPROUT} \cite{somerstep2025}.

Future research should incorporate standardised comparison baselines:   \textit{ (i)  random routing, (ii) the most cost-effective LLM with the highest performance for each query (oracle routing), (iii) ) the stand-alone best-performing LLM, and (iv) alternative routing strategies reviewed in this survey.} Such comparisons will determine whether the efficacy of routing derives from the available routing candidates or the routing architecture. The performance gap between the best stand-alone LLM and gold standard routing quantifies the theoretical margin for improvement.  Direct comparison of stand-alone LLMs with routing systems will quantify the added value of routing architectures across benchmark datasets.

\subsubsection{Using complementary routing options}\label{Using complementary routing options}

Routing to candidates with complementary rather than redundant skills is essential to optimise routing performance. For example, multiple models of the same size (e.g., 7B parameters) trained on a generalist corpus may exhibit some complementarity due to differences in their training datasets. However, we cannot expect significant performance enhancements for queries on specific or complex topics requiring advanced reasoning. The primary objective of routing is to maximise quality. Incorporating models with varying parameter sizes or trained on specialised domains allows the routing strategy to adapt effectively to various contexts. In addition, when evaluating different routing approaches, it is important to consider how complementary and efficient the different routing candidates are. 

\subsubsection{Consider all steps in the LLM-based system as routing possibilities}\label{Consider all steps in the LLM-based system as routing possibilities}

Most studies focused on the generation step by selecting the most appropriate LLM to answer the user query. However, current LLM-based systems like conversational agents typically include several additional steps. The routing approach can be effectively applied during the embedding step in the RAG architecture. This includes routing between different embedding strategies, such as dense or sparse vectors, or the selection of fine-tuned embedding models \cite{gao2023}. It also allows routing to databases tailored to specific topics, selecting appropriate similarity functions such as cosine similarity or BM25, and selecting the most appropriate prompting approaches \cite{gao2023}. This framework can also be extended to facilitate routing between static knowledge sources, such as databases, and dynamic knowledge sources, such as website searches. Some authors have studied the use of routing for additional steps, such as context size selection \cite{li2024}, prompt strategies \cite{ning2024}, and even alternative pipeline designs \cite{jeong2024}. Viewing these systems as dynamic rather than traditional static models facilitates optimisation at each stage and enhances modularity \cite{gao2024}. This transforms query answering into a singular dynamic process that depends on the specific query being addressed.

\subsubsection{Towards autonomous adaptive routing strategies}\label{Towards autonomous adaptive routing strategies}

A significant limitation of existing routing systems is that they are static. Introducing new routing candidates, whether models, prompts, workflows or modules, typically requires the entire routing infrastructure to be retrained. This inflexibility hinders scalability and adaptability in dynamic environments. 
Recent research suggests promising alternatives that support the addition of new components without the need for retraining. These strategies typically involve projecting routing options into a shared semantic space, which allows for generalisation to configurations that have not been seen before \cite{feng2024, jitkrittum2025, li2025llmbandit}. Viewing the router as an autonomous, adaptive agent enables continuous alignment with environmental changes, resource availability and evolving objectives. 

\section{Conclusion}\label{Conclusion}
Routing in an LLM-based system can be defined as a process that aims to select components of the system that maximise performance while minimising cost. The definition of the cost to be minimised and the score function to be maximised is essential to constructing a routing algorithm. We classified routing strategies as pre-generation or post-generation. We highlighted that implementing a routing strategy to maintain performance while minimising cost can be efficient and does not necessarily require high resources. We highlighted the need to develop products based on the lightweight strategies discussed in this survey. We emphasised the need to work on well-designed benchmarks across routing strategies to assess which approach offers the best potential by proposing key baseline comparisons. We also discussed the importance of considering computational and environmental costs in addition to financial costs. Finally, we explore future perspectives for routing improvement through the complementarity of routing options and its potential confounding effect. We consider LLM-based systems as dynamic systems and highlight the need for the router to be able to autonomously generalise to new routing options.

\break

\printbibliography

 \break
\appendix

\section{Appendix A - Descriptive table of the routing strategies}\label{annex2}

\begin{table*}[!htbp]
\centering
\small
  \caption{A description of the various works included in the survey}
  \label{tab:tab_annex}
\makebox[\textwidth][c]{%
\begin{tabular}{p{0.15\textwidth}p{0.20\textwidth}p{0.23\textwidth}p{0.20\textwidth}p{0.07\textwidth}p{0.20\textwidth}}
    \toprule
    \textbf{Study} & \textbf{Task} & \textbf{Benchmarks} & \textbf{Compared Strategies} & \textbf{Routing Step} & \textbf{\mbox{Routing Approach}} \\
    \midrule
\citet{aggarwal2024} & Short and Multi-Choice Q/A, Text understanding, Reasoning & QASPER,QUALITY, COQA, MUTUAL, DIPLOMAT & Stand-alone models, FrugalGPT \cite{chen2023}, HybridLLM (\cite{ding2024}) & Post & Repeated calls \\ 
    \midrule
\citet{chai2024} & Multi-domain Q/A & MMLU Expert & Prompting methods, gold routing, LLM-Blender & Pre & LLM fine-tuning\\ 
    \midrule
\citet{chen2023} & Short Q/A, Text classification & HEADLINES, OVERRULING, COQA & Stand-alone LLM, top LLM & Post & Answer confidence inference\\
    \midrule
\citet{chen2024routerdc} & Multi-domain and Multi-Choice Q/A, Code generation & MMLU, GSM8K, CMMLU, ARC-Challenge, HumanEval, PreAlgebra, MBPP, C-EVAL & Stand-alone models, majority voting, ZOOTER \cite{lu2024}, multi-class classification and clustering & Pre & Query similarity\\
    \midrule
\citet{chuang2024} & Multi-domain and Multi-Choice Q/A, Code generation & MMLU, OpenbookQA, GSM8K, MedQA & Verbalised confidence, logits-based uncertainty, random routing & Post & LLM fine-tuning \\
    \midrule
\citet{dekoninck2024} (1) & Multi-domain and Multi-Choice Q/A & ARC-Challenge, MMLU-Pro, MixEval, GSM8k  & Linear interpolation, linear optimisation programs (query-based classification, cascading) & Post & Answer confidence inference \\
    \midrule
\citet{dekoninck2024} (2) & Multi-domain and Multi-Choice Q/A & ARC-Challenge, MMLU-Pro, MixEval, GSM8k  & Linear interpolation, linear optimisation programs (query-based classification, cascading) & Pre & Query complexity inference \\
    \midrule
\citet{ding2024} & Instructions & MixInstruct & Random routing, smallest LLM, largest LLM & Pre & Query complexity inference \\
    \midrule
\citet{feng2024} & Multi-domain Q/A, Text understanding, Summarization & Alpaca, GSM8K, SQUAD, Multi-News & FrugalGPT \cite{chen2023}, HybridLLM \cite{ding2024}, prompt routing, smallest LLM, largest LLM, gold routing, bandit-based model & Pre & Knowledge Graph \\
    \midrule
\citet{hari2023} & Multi-Domain Text Corpus & Expert corpus (e.g., Pile-CC, Pubmed Central, ArXiv ) & Stand-alone models, Gorilla (\cite{patil2023}) & Pre & Reward-based inference \\
    \midrule
\citet{hu2024} & Multi-domain and Multi-Choice Q/A, Instructions, Reasoning & MMLU, MT-Bench, MBPP, HellaSwag, Winogrande, GSM8K, Arc-Challenge & Stand-alone models, K-NN, MLP, linear interpolation, oracle routing & Pre & Query complexity inference \\
   \bottomrule
  \end{tabular}}
\end{table*}

\begin{table*}[!htbp]
\centering
\small
\makebox[\textwidth][c]{%
\begin{tabular}{p{0.15\textwidth}p{0.20\textwidth}p{0.23\textwidth}p{0.20\textwidth}p{0.07\textwidth}p{0.20\textwidth}}
    \toprule
    \textbf{Study} & \textbf{Task} & \textbf{Benchmarks} & \textbf{Compared Strategies} & \textbf{Routing Step} & \textbf{\mbox{Routing Approach}} \\
    \midrule

\citet{jain2024} & Multi-domain Q/A &  MMLU Pro, GSM8K & Stand-alone models  & Pre & Domain classification\\

\citet{jang2023} & Q/A, Reasoning, Text understanding, Text classification, Instructions & RTE, CB, ANLI, COPA, HellaSWAG, Storycloze, WinoGrande, WSC, WiC, Big-Bench, wiki-auto, HGen, COVID-QA, ELI5 & Stand-alone models  & Pre & Query similarity\\
    \midrule
\citet{jeong2024} & Q/A, Text understanding & Single-step Q/A (e.G., SQuAD-v1.1, TriviaQA), Multi-step Q/A (e.g., MuSiQue, HotpotQA)& Adaptive
Retrieval, Self-RAG, gold routing & Pre & Query complexity inference \\
    \midrule
\citet{jitkrittum2025} & Multi-Domain and Multi-Choice Q/A, Code generation, Reasoning, Instructions & EmbedLLM, RouterBench, Mix-Instruct & Random, KNN & Pre & Clustering previous interactions\\
    \midrule
\citet{lee2024} (1) & DST & MultiWOZ, SGD & Prompt-DST, IC-DST and DS2-T5  & Pre & Query similarity\\ 
    \midrule
\citet{lee2024} (2) & DST & MultiWOZ, SGD & Prompt-DST, IC-DST and DS2-T5  & Post & Sequence probability \\ 
    \midrule
\citet{lee2025} & Adversarial attack detection & HarmBench, OpenAI Moderation, ToxicChat, WildGuardMix, XSTest
 &  Uncertainty (e.g. entropy) & Pre & Query complexity inference \\ 
 \midrule
\citet{li2024} & Q/A, Fact extraction, Text understanding & NarrativeQA, QASPER, MultiFieldQA, HotpotQA,etc. & Naïve RAG, long context  & Post & Prompt-based routing\\ 
\midrule
\citet{liu2024meswitch} & Multi-domain and Multi-choice Q/A, Code generation & MMLU, GSM8K, MATH, HumanEval, MBPP, C-Eval and C-MMLU & Stand-alone models  & Pre & LLM fine-tuning \\
\midrule
\citet{lu2024} & Q/A, Instructions,  Multi-domain Q/A, Code generation & AlpacaEval, FLASK, MT-Bench, MMLU, GSM8K, HumanEval & Stand-alone models, overall best LLM, oracle routing  & Pre & Reward-based inference \\
\midrule
\citet{malekpour2024} & Text-to-SQL & BIRD & Stand-alone models  & Pre & Query similarity, Query complexty inference \\ 
\midrule
\citet{manias2024} & Intent detection & Custom Dataset & /  & Pre & Query similarity \\ 
\midrule
\citet{mohammadshahi2024} & Multi-Domain Q/A & MMLU & Stand-alone models & Pre & Decoder-based encoding \\
\midrule
\citet{nguyen2024} & Text classification & IMDB, SST-2 & Stand-alone models  & Pre & State-based algorithms\\
\midrule
\citet{ning2024} & Q/A & FastChat, LLMZoo & No routing  & Pre & Query complexity inference\\ 
    \bottomrule
  \end{tabular}}
\end{table*}

\begin{table*}[!htbp]
\centering
\small
\makebox[\textwidth][c]{%
\begin{tabular}{p{0.15\textwidth}p{0.20\textwidth}p{0.23\textwidth}p{0.20\textwidth}p{0.07\textwidth}p{0.20\textwidth}}
    \toprule
    \textbf{Study} & \textbf{Task} & \textbf{Benchmarks} & \textbf{Compared Strategies} & \textbf{Routing Step} & \textbf{\mbox{Routing Approach}} \\
    \midrule

\citet{ning2024} & Q/A & FastChat, LLMZoo & No routing  & Pre & Prompt-based routing\\ 
\midrule
\citet{ong2024} & Multi-Domain Q/A, Instructions & MT Bench, MMLU, GSM8K & Random routing  & Pre & Query complexity inference, preference similarity, recommendation system, LLM fine-tuning \\ 
\midrule
\citet{patil2023} & Code generation & Custom API calling dataset & Stand-alone models   & Pre & LLM fine-tuning \\ 
\midrule
\citet{pichlmeier2024} & Feasibility & / & /  & Pre & Clustering previous interactions\\
\midrule
\citet{ramirez2024} & Text classification, Multi-Domain and Multi-Choice Q/A, Fact Extraction & Wikifact, Openbook, ISEAR, FEVER, bAbI, Natural Questions, SST-2, CR, RT-Polarity & FrugalGPT \cite{chen2023}, HybridLLM \cite{ding2024}, query-based classification  & Post & Token probability\\
\midrule
\citet{sakota_2024} & Text classification, Multi-Domain and Multi-Choice Q/A & MMLU, GSM8K, WikiFact, RAFT, LegalSupport, etc. & Oracle routing, stand-alone models   & Pre & Query complexity inference \\ 
\midrule
\citet{shen2023} & Task Planning & Custom Dataset & /  & Pre & Prompt-based routing \\
\midrule
\citet{shnitzer2023} & Multi-Domain Q/A, Instructions, Text classification & HELM, MixInstruct & Stand-alone models, overall best LLM, oracle routing,  & Pre & Query complexity inference\\
\midrule
\citet{sikeridis2024} & Q/A & HC3 & Random routing, stand-alone models  & Pre & Stateless algorithms \\
\midrule
\citet{simonds2024} & Multi-Domain and Multi-Choice Q/A, Code generation & MMLU, MMLU Pro, GPQA, HumanEval, College Math, MATH, GSM8k, Olympiad Bench & Stand-alone models  & Pre & Domain Classification\\
\midrule
\citet{somerstep2025} & Multi-Domain and Multi-Choice Q/A, Code generation, Instructions, Reasoning & IFEval, GSM8K, MMLU, Arc-Challenge, MBPP, MT-Bench, WinoGrande, HellaSWAG, MMLU-Pro, GPQA, MATH, BigBenchHard, MuSR, OpenHermes-2.5, RAGBench &  Random forests, RouteLLM \cite{ong2024}  & Pre & Query complexity inference\\
\midrule
\citet{srivatsa2024} & Multi-domain Q/A & MMLU, GSM8K & Stand-alone models, random routing, oracle routing   & Pre & Clustering previous interactions, query complexity inference\\
\midrule
\citet{stripelis2024} & Multi-domain Q/A, Code generation & Ai2-ARC, GSM8K, MBPP, PubMedQA & Random routing, stand-alone models, oracle routing & Pre & Query similarity, Query complexity inference\\
    \bottomrule
  \end{tabular}}
\end{table*}

\begin{table*}[!htbp]
\centering
\small
\makebox[\textwidth][c]{%
\begin{tabular}{p{0.13\textwidth}p{0.20\textwidth}p{0.23\textwidth}p{0.20\textwidth}p{0.07\textwidth}p{0.20\textwidth}}
    \toprule
    \textbf{Study} & \textbf{Task} & \textbf{Benchmarks} & \textbf{Compared Strategies} & \textbf{Routing Step} & \textbf{\mbox{Routing Approach}} \\
    \midrule
\citet{wang2024CoE} & Multi-domain and Multi-Choice Q/A, Code generation, Reasoning & MMLU Pro, GSM8K, Winogrande, Big Bench Hard, MMMU, MMStar & Stand-alone models & Pre & Domain classification, Query complexity inference\\
\midrule
\citet{wang2025mixllm} & Multi-domain and Multi-Choice Q/A, Code generation, Reasoning, Instructions & GSM8K, MMLU, Arc-Challenge, MBPP, MT-Bench, WinoGrande, HellaSWAG & Stand-alone models, oracle routing, Random, RouteLLM \cite{ong2024}, Zooter, \cite{lu2024} RouterBench \cite{hu2024}, OptLLM \cite{liu2024optllm}, MetaLLM \cite{nguyen2024}  & Pre & Query complexity inference\\
\midrule
\citet{yue2024} & Multi-Domain Q/A, Text classification, Reasoning, Text understanding & GSM8K, ASDIV, TabMWP, Big-Bench Hard, CREPE & Stand-alone models  & Pre & Query complexity inference \\
\midrule
\citet{zhang2023} & Instructions & ToolBench & Stand-alone models & Post & Code execution\\ 
\midrule
\citet{zhang2025} & Multi-Choice and Multi-Domain Q/A & Natural Questions, TriviaQA, MAWPS, POPQA & TensorOpera \cite{stripelis2024}, RouteLLM \cite{ong2024}  & Pre & Recommendation system \\ 
\midrule
\citet{zhao2024} & Multi-Domain and Multi-Choice Q/A, Code generation , Instructions & MMLU, Hellaswag, GSM8K, ARC-Challenge, Winogrande, MBPP, MT-Bench  & Linear SVM, KNN and MLP  & Pre & Preference similarity\\
\midrule
\citet{zhuang2025} & Multi-Choice and Multi-Domain Q/A, Code generation & MMLU, GSM8K, GPQA, TruthfulQA, ASDIV, LogiQA, MathQA, MedMCQA, PIQA, SocialQA & Best Model & Pre & Recommendation system \\
    \bottomrule
  \end{tabular}}
\end{table*}

\end{document}